\documentclass{article}

\usepackage[T1]{fontenc}
\usepackage{tgtermes}
\usepackage{amsmath}
\usepackage{enumitem}

\usepackage[scaled=0.92]{PTSans}
\usepackage{amssymb}

\usepackage{scalefnt,letltxmacro}
\LetLtxMacro{\oldtextsc}{\textsc}
\usepackage[ttdefault=true]{AnonymousPro}

\usepackage[usenames,dvipsnames]{xcolor}
\definecolor{shadecolor}{gray}{0.9}

\usepackage[english]{babel}
\usepackage[parfill]{parskip}
\usepackage{afterpage}
\usepackage{framed}

{\endMakeFramed}

\DeclareRobustCommand{\parhead}[1]{\textbf{#1}~}

\usepackage{lineno}

\usepackage{ragged2e}

\newcounter{parcount}

\usepackage{graphicx}
\usepackage[labelfont=bf]{caption}
\usepackage[format=hang]{subcaption}

\usepackage{booktabs}

\usepackage[algoruled]{algorithm2e} 
\usepackage{listings}
\usepackage{fancyvrb}
\fvset{fontsize=\normalsize}

\usepackage{natbib}

\usepackage[colorlinks,linktoc=all]{hyperref}
\hypersetup{citecolor=Blue}
\hypersetup{linkcolor=MidnightBlue}
\hypersetup{urlcolor=MidnightBlue}

\newcommand{\myeqp}[1]{Eq.~\ref{eq:#1}}
\newcommand{\mysec}[1]{Section~\ref{sec:#1}}
\newcommand{\mytable}[1]{Table~\ref{table:#1}}
\newcommand{\myfig}[1]{Figure~\ref{fig:#1}}

\usepackage[acronym,smallcaps,nowarn]{glossaries}
\glsdisablehyper{}

\newcommand{\red}[1]{\textcolor{BrickRed}{#1}}

\newcommand{\blue}[1]{\textcolor{MidnightBlue}{#1}}

\lstdefinestyle{mystyle}{
    commentstyle=\color{OliveGreen},
    keywordstyle=\color{BurntOrange},
    numberstyle=\tiny\color{black!60},
    stringstyle=\color{MidnightBlue},
    basicstyle=\ttfamily,
    breakatwhitespace=false,
    breaklines=true,
    captionpos=b,
    keepspaces=true,
    numbers=left,
    numbersep=5pt,
    showspaces=false,
    showstringspaces=false,
    showtabs=false,
    tabsize=2
}
\newacronym{SVI}{svi}{stochastic variational inference}
\newacronym{ELBO}{elbo}{Evidence Lower BOund}
\newacronym{DEF}{def}{deep exponential family}
\newacronym{SBN}{sbn}{sigmoid belief network}

\newacronym{DLGM}{dlgm}{deep latent Gaussian model}

\newacronym{MF}{mf}{mean-field}
\newacronym{HVM}{hvm}{hierarchical variational model}
\newacronym{COVI}{compound vi}{compound variational inference}
\newacronym{VGP}{variational gp}{variational Gaussian process}
\newacronym{BBVI}{bbvi}{black box variational inference}
\newacronym{ADVI}{advi}{automatic differentiation variational inference}
\newacronym{NF}{nf}{normalizing flows}
\newacronym{CVI}{copula vi}{copula variational inference}
\newacronym{VAE}{vae}{variational autoencoder}
\newacronym{IWAE}{iwae}{importance weighted autoencoder}
\newacronym{NVIL}{nvil}{neural variational inference}
\newacronym{MIXTURE}{mixture}{}
\newacronym{DSVI}{dsvi}{}

\newacronym{GP}{gp}{Gaussian process}
\newacronym{GPLVM}{gplvm}{Gaussian process latent variable model}
\newacronym{ARD}{ard}{automatic relevance determination}
\newacronym{SVIGP}{svigp}{}
\newacronym{KISSGP}{kiss-gp}{}
\newacronym{FASTFOOD}{fastfood}{}

\glsunset{ADVI}
\glsunset{SVIGP}
\glsunset{MIXTURE}
\glsunset{CVI}
\glsunset{DSVI}
\glsunset{FASTFOOD}
\glsunset{KISSGP}

\DeclareRobustCommand{\mb}[1]{\ensuremath{\boldsymbol{\mathbf{#1}}}}

\newcommand{\KL}{\textrm{KL}}

\newcommand{\E}{\mathbb{E}}

\newcommand{\cL}{\mathcal{L}}
\newcommand{\g}{\, | \,}

\newcommand{\expfam}{\textsc{expfam}}

\newcommand{\mbz}{\mb{z}}
\newcommand{\mbx}{\mb{x}}
\newcommand{\mbu}{\mb{u}}
\newcommand{\mbW}{{\mb{W}}}
\newcommand{\mblambda}{{\mb{\lambda}}}
\newcommand{\mbtheta}{{\mb{\theta}}}
\newcommand{\mbphi}{\mb{\phi}}
\newcommand{\mbepsilon}{\mb{\epsilon}}
\newcommand{\mbpi}{{\mb{\pi}}}
\newcommand{\mbmu}{{\mb{\mu}}}

\newcommand{\qmf}{{q_{\textsc{mf}}}}

\newcommand{\qhvm}{{q_{\textsc{hvm}}}}
\newcommand{\qpost}{{q_{\textsc{post}}}}

\newcommand{\cLmf}{{\cL_{\textsc{mf}}}}

\renewcommand{\d}[1]{\ensuremath{\operatorname{d}\!{#1}}}

\usepackage{tikz}
\usetikzlibrary{bayesnet}

\pgfdeclarelayer{edgelayer}
\pgfdeclarelayer{nodelayer}
\pgfsetlayers{edgelayer,nodelayer,main}

\definecolor{hexcolor0xbfbfbf}{rgb}{0.749,0.749,0.749}

\tikzset{>=latex}
\tikzstyle{none}   = [inner sep=0pt]

\tikzstyle{line}  = [ - ]
\tikzstyle{arrow}  = [ ->, shorten <=1pt, shorten >=1pt ]
\tikzstyle{ardash} = [ dotted, ->, shorten <=1pt, shorten >=1pt ]

\tikzstyle{empty}=[circle,opacity=0.0,text opacity=1.0,inner sep=0pt,minimum
width=0pt,minimum height=0pt]
\tikzstyle{box}=[rectangle,fill=White,draw=Black]
\tikzstyle{filled}=[circle,fill=hexcolor0xbfbfbf,draw=Black]
\tikzstyle{hollow}=[circle,fill=White,draw=Black]
\tikzstyle{param}=[rectangle,fill=Black,draw=Black,inner sep=0pt,minimum width=4pt,minimum height=4pt]

\usepackage{pifont}
\newcommand{\cmark}{\ding{51}}%
\newcommand{\xmark}{\ding{55}}%
\usepackage[accepted]{icml2016}
\usepackage{times}

\icmltitlerunning{Hierarchical Variational Models}

\begin{document}

\twocolumn[
\icmltitle{Hierarchical Variational Models}

\icmlauthor{Rajesh Ranganath}{rajeshr@cs.princeton.edu}
\icmladdress{Princeton University, 35 Olden St., Princeton, NJ 08540}
\icmlauthor{Dustin Tran}{dustin@cs.columbia.edu}
\icmlauthor{David M.~Blei}{david.blei@columbia.edu}
\icmladdress{Columbia University, 500 W 120th St., New York, NY 10027}

\vskip 0.3in
]

\begin{abstract}
\vskip 0.1in
  Black box variational inference allows researchers to easily prototype and
  evaluate an array of models. Recent advances allow such algorithms to scale to high dimensions.
  However, a central question remains: How to specify an expressive
  variational distribution that maintains efficient computation?  To
  address this, we develop \glspl{HVM}. \acrshortpl{HVM}
  augment a variational approximation with a prior on its
  parameters, which allows it to capture complex structure for both
  discrete and continuous latent variables.
  The algorithm we develop is black box, can be used
  for any \acrshort{HVM}, and has the same computational efficiency as
  the original approximation.
  We study \acrshortpl{HVM} on a variety of
  deep discrete latent variable models. \acrshortpl{HVM} generalize
  other expressive variational distributions and maintains higher
  fidelity to the posterior.
\end{abstract}

\section{Introduction}

\Gls{BBVI} is important to realizing the potential of modern applied
Bayesian statistics.  The promise of \gls{BBVI} is that an
investigator can specify any probabilistic model of hidden and
observed variables, and then efficiently approximate its posterior
without additional effort~\citep{Ranganath:2014}.

\gls{BBVI} is a form of variational
inference~\citep{Jordan:1999}.  It sets up a
parameterized family of distributions over the latent variables and
then optimizes the parameters to be close to the posterior. Most applications of variational
inference use the mean-field family. Each variable is
independent and governed by its own parameters.

Though it enables efficient inference, the mean-field family is
limited by its strong factorization. It cannot capture posterior
dependencies between latent variables, dependencies which both
improve the fidelity of the approximation and are sometimes of intrinsic
interest.

To this end, we develop \glsreset{HVM}\emph{\glspl{HVM}}, a class of families that goes beyond the mean-field and,
indeed, beyond directly parameterized variational families in general.
The main idea behind our method is to treat the variational family as
a model of the latent variables and then to expand this model
hierarchically.  Just as hierarchical Bayesian models induce
dependencies between data, hierarchical variational models induce
dependencies between latent variables.

We develop an algorithm for fitting \glspl{HVM} in the context of
black box inference. Our algorithm is as general and computationally
efficient as \gls{BBVI} with the mean-field family, but it finds
better approximations to the posterior.  We demonstrate \glspl{HVM}
with a study of approximate posteriors for several variants of deep
exponential families~\citep{Ranganath:2015}; \glspl{HVM} generally
outperform mean-field variational inference.

\parhead{Technical summary.} Consider a posterior distribution
$p(\mbz\g\mbx)$, a distribution on $d$ latent variables
$\mbz_1,\ldots,\mbz_d$ conditioned on a set of observations $\mbx$.
The mean-field family is a factorized distribution of the latent
variables,
\begin{align}
  \label{eq:mean-field}
  \qmf(\mbz;\mblambda) = \textstyle \prod_{i=1}^d q(z_i ;
  \mblambda_i).
\end{align}
We fit its parameters $\mblambda$ to find a variational
distribution that is close to the exact posterior.

By positing \myeqp{mean-field} as a model of the latent variables, we
can expand it by placing a prior on its parameters.  The result is a
\textit{\acrlong{HVM}}, a two-level distribution that first
draws variational parameters from a prior $q(\mblambda ; \mbtheta)$
and then draws latent variables from the corresponding likelihood
(\myeqp{mean-field}). \glspl{HVM} induce a family that marginalizes out the
mean-field parameters,
\begin{align}
  \label{eq:mixed-mf}
  \qhvm(\mbz ; \mbtheta)
  &= \int q(\mblambda ; \mbtheta) \prod_i q(z_i \g \mblambda_i)
    \d\mblambda.
\end{align}
This expanded family can capture both posterior dependencies between the
latent variables and more complex marginal distributions, thus better
inferring the posterior.  (We note that
during inference the variational ``posterior''
$q(\mblambda \g \mbz, \mbtheta)$ will also play a role; it is the
conditional distribution of the variational parameters given a
realization of the hidden variables.)

Fitting an \gls{HVM} involves optimizing the variational hyperparameters
$\mbtheta$, and our algorithms for solving this problem maintain
the computational efficiency of \gls{BBVI}. Note the prior is a
choice.  As one example, we use mixture models as a prior of the
mean-field parameters.  As another, we use normalizing
flows~\citep{Rezende:2015}, expanding their scope to a broad class of
non-differentiable models.

\section{Hierarchical Variational Models}

Recall, $p(\mbz\g\mbx)$ is the posterior.
Variational inference frames posterior inference as
optimization: posit a family of distributions $q(\mbz; \mblambda)$,
parameterized by $\mblambda$, and minimize the KL divergence to the
posterior distribution~\citep{Jordan:1999,Wainwright:2008}.

Classically, variational inference uses the mean-field family. In the
mean-field family, each latent variable is assumed independent and
governed by its own variational parameter (\myeqp{mean-field}).  This
leads to a computationally efficient optimization problem that can be
solved (up to a local optimum) with coordinate
descent~\citep{Bishop:2006,Ghahramani:2001} or gradient-based
methods~\citep{Hoffman:2013,Ranganath:2014}.

Though effective, the mean-field factorization compromises the
expressiveness of the variational family: it abandons any
dependence structure in the posterior, and it cannot in general
capture all marginal information.  One of the challenges of
variational inference is to construct richer approximating families---thus yielding high
fidelity posterior approximations---and while still being
computationally tractable. We develop a framework
for such families.

\subsection{\Acrlongpl{HVM}}

Our central idea is to draw an analogy between probability models of
data and variational distributions of latent variables.  A probability
model outlines a family of distributions over data, and how large that
family is depends on the model's complexity. One common approach to
expanding the complexity, especially in Bayesian statistics, is to
expand a model hierarchically, i.e., by placing a prior on the
parameters of the likelihood.  Expanding a model hierarchically has
distinct advantages: it induces new dependencies between the data,
either through shrinkage or an explicitly correlated
prior~\citep{efron2012large}, and it enables
us to reuse algorithms for the simpler model within algorithms for the
richer model~\citep{Gelman:2007}.

We use the same idea to expand the complexity of the mean-field
variational family and to construct \glsreset{HVM}\emph{\glspl{HVM}}.  First, we
view the mean-field family of~\myeqp{mean-field} as a simple model of
the latent variables.  Next, we expand it hierarchically.  We
introduce a ``{variational prior}'' $q(\mblambda ;\mbtheta)$ with
``{variational hyperparameters}'' $\mbtheta$ and place it on the
mean-field model (a type of ``{variational likelihood}'').
Marginalizing out the prior gives $\qhvm(\mbz ; \mbtheta)$, the
hierarchical family of distributions over the latent variables in
\myeqp{mixed-mf}. This family enjoys the advantages of
hierarchical modeling in the context of variational inference: it
induces dependence among the latent variables and allows us to reuse
simpler computation when fitting the more complex family.

\begin{figure}[tb]
\begin{subfigure}[t]{0.45\columnwidth}
  \centering
  \begin{tikzpicture}[x=1.7cm,y=1.8cm]

  \node[obs]                     (z1)      {$z_1$} ;
  \node[obs, right=of z1, xshift=-1.25cm]        (z2)      {$z_2$} ;
  \node[obs, right=of z2, xshift=-1.25cm]        (z3)      {$z_3$} ;
  \factor[above=of z1, yshift=0.75cm]  {lambda1} {$\mblambda_1$} {} {};
  \factor[above=of z2, yshift=0.75cm]      {lambda2} {$\mblambda_2$} {} {};
  \factor[above=of z3, yshift=0.75cm]  {lambda3} {$\mblambda_3$} {} {};

  \edge{lambda1}{z1};
  \edge{lambda2}{z2};
  \edge{lambda3}{z3};

  \plate[inner sep=0.40cm, yshift=0.35cm,
    label={[xshift=-14pt,yshift=14pt]south east:$n$}] {plate1} {
    (z1)(z2)(z3)(lambda1)(lambda2)(lambda3)
  } {};

  \node[below=of z2, yshift=0.25cm] (label)
  {{\small \textbf{(a)}} \textsc{mean-field model}} ;

\end{tikzpicture}
  \label{sub:mean_field}
\end{subfigure}
\hspace{1em}
\begin{subfigure}[t]{0.42\columnwidth}
  \centering
  \begin{tikzpicture}[x=1.7cm,y=1.8cm]

  \node[obs]                     (z1)      {$z_1$} ;
  \node[obs, right=of z1, xshift=-1.25cm]        (z2)      {$z_2$} ;
  \node[obs, right=of z2, xshift=-1.25cm]        (z3)      {$z_3$} ;
  \node[latent, above=of z1, yshift=-0.5cm]  (lambda1) {\small$\mblambda_1$} {} {};
  \node[latent, above=of z2, yshift=-0.5cm]      (lambda2) {\small$\mblambda_2$} {} {};
  \node[latent, above=of z3, yshift=-0.5cm]  (lambda3) {\small$\mblambda_3$} {} {};

  \factor[above=of lambda2] {theta} {$\mbtheta$} {} {};

  \edge{lambda1}{z1};
  \edge{lambda2}{z2};
  \edge{lambda3}{z3};
  \edge{theta}{lambda1};
  \edge{theta}{lambda2};
  \edge{theta}{lambda3};
  \draw[-] (lambda1) -- (lambda2) -- (lambda3);

  \plate[inner sep=0.35cm, yshift=0.20cm,
    label={[xshift=-14pt,yshift=14pt]south east:$n$}] {plate1} {
    (z1)(z2)(z3)(lambda1)(lambda2)(lambda3)
  } {};

  \node[below=of z2, yshift=0.25cm] (label)
  {{\small \textbf{(b)}} \textsc{hierarchical model}} ;

\end{tikzpicture}
  \label{sub:exvm}
\end{subfigure}
\vspace{-2ex}
\caption{Graphical model representation. \textbf{(a)} In mean-field models, the
latent variables are strictly independent. \textbf{(b)} In hierarchical
variational
models, the latent variables are governed by a prior distribution on their
parameters, which induces arbitrarily complex structure.
\vspace{-5ex}
}
\label{fig:exvm}
\end{figure}
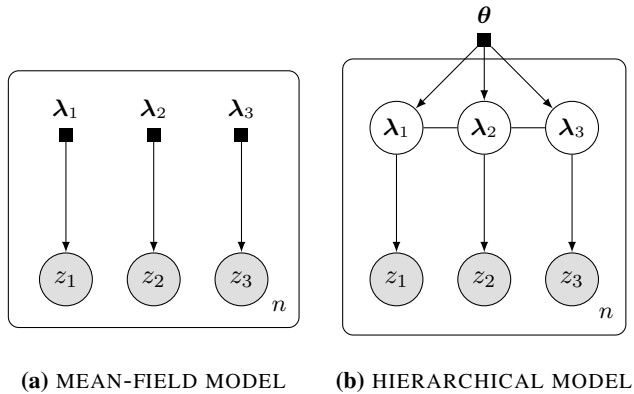

\myfig{exvm} illustrates the difference between the mean-field family
and an \gls{HVM}.  Mean-field inference fits the variational
parameters $\{\mblambda_1, \ldots, \mblambda_d\}$ so that the
factorized distribution is close to the exact posterior; this
tries to match the posterior marginal for each variable.
Using the same principle, \gls{HVM} inference fits the variational
hyperparameters so $\qhvm(\mbz ; \mbtheta)$ is close to the exact
posterior.  This goes beyond matching marginals because of the
shrinkage effects among the variables.

\begin{figure}[!tb]
\begin{subfigure}[t]{0.49\columnwidth}
  \centering
  \includegraphics[width=\columnwidth]{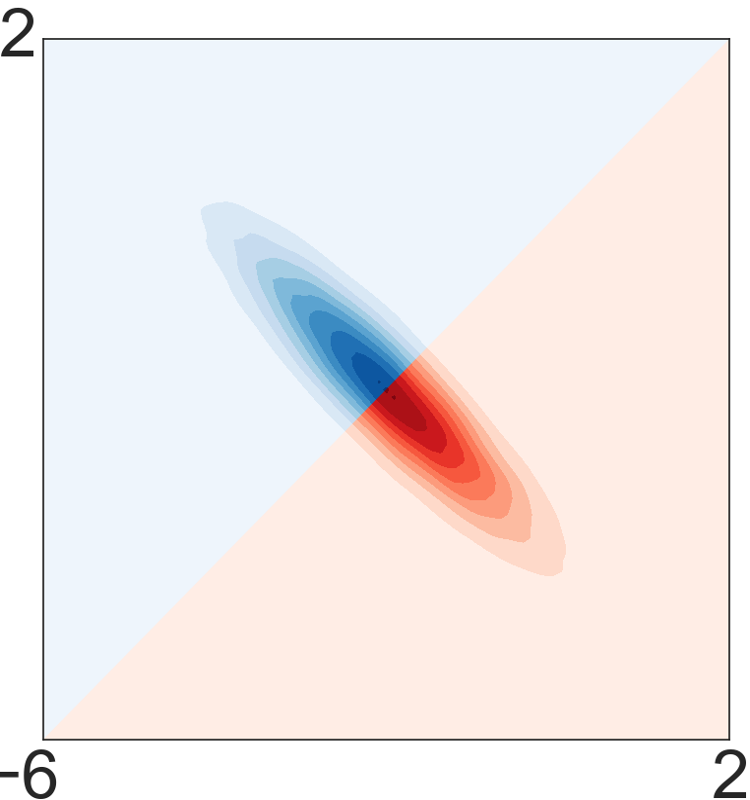}
  \label{sub:gaussian}
  \vspace{-3ex}
  \caption{$\operatorname{Normal}(\mblambda;\mbtheta)$}
\end{subfigure}
\hspace{-0.5em}
\begin{subfigure}[t]{0.49\columnwidth}
  \centering
  \includegraphics[width=0.96\columnwidth]{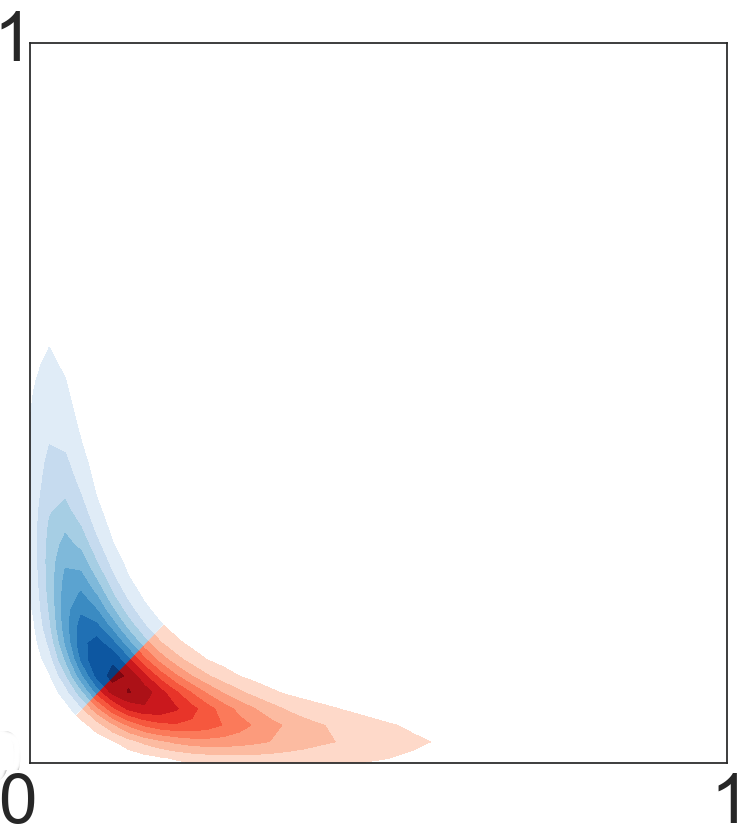}
  \label{sub:gamma}
  \vspace{-0.5ex}
  \caption{$\prod_{i=1}^2
  \operatorname{Gamma}(z_i\g\mblambda_i)$}
\end{subfigure}
\vspace{-1ex}
\caption{%
\textbf{(a)} $q(\mblambda;\mbtheta)$: The (reparameterized) natural
parameters assume a multivariate prior, with different
areas indicated by \red{red} and \blue{blue}.
\textbf{(b)} $\prod_{i=1}^2 q(z_i\g\mblambda_i)$: The latent variables are drawn from a mean-field
family, colorized according to its drawn parameters' color.
\vspace{-2ex}
}
\label{fig:gaussian_gamma}
\end{figure}

\myfig{gaussian_gamma} is a simple example. The variational
family posits each $z_i$ as a scalar from an exponential family. The
variational
parameters $\mblambda_i$ are the corresponding natural parameters, which
are unconstrained. Now place a Gaussian
prior on the mean-field parameters, with a full covariance matrix.
The resulting \gls{HVM} is a two-level
distribution: first draw the complete set of variational parameters
$\{\mblambda_1, \ldots, \mblambda_d\}$ from a Gaussian
(\myfig{gaussian_gamma}a); then draw each
$z_i$ from its corresponding natural parameter
(\myfig{gaussian_gamma}b).  The covariance on the
variational parameters induces dependence among the
$z_i$'s, and the marginal of each $z_i$ is an integrated likelihood; thus
this \gls{HVM} is more flexible than the mean-field family.

In general, if the \gls{HVM} can capture the same marginals then
$\qhvm(\mbz ; \mbtheta)$ is more expressive than the mean-field
family.\footnote{%
Using an HVM to ``regularize'' the variational family, i.e., to induce
dependence but limit the marginals, is an interesting avenue for
future work.  In the appendix, we relate \acrshortpl{HVM} to empirical
Bayes and methods in reinforcement learning.}
As in the example, the \gls{HVM} induces dependence
among variables and also expands the family of possible
marginals that it can capture. In \mysec{optimization} we see that,
even with this more expressive family, we can develop a black box
algorithm for \glspl{HVM}. It exploits the mean-field structure of the
variational likelihood and enjoys the corresponding computational
advantages. First, we discuss how to specify an \gls{HVM}.

\subsection{Specifying an HVM}
\label{sec:specifying}

We can construct an \acrshort{HVM} by placing a prior on any existing
variational approximation. An \gls{HVM} has two components: the
variational likelihood $q(\mbz\mid\mblambda)$ and the prior
$q(\mblambda;\mbtheta)$.  The likelihood comes from a variational
family that admits gradients; here we focus on the mean-field family.
As for the prior, the distribution of
$\{\mblambda_1, \ldots, \mblambda_d\}$ should not have the same
factorization structure as the variational likelihood---otherwise it
will not induce dependence between latent variables. We outline
several examples of variational priors.

\parhead{Variational prior: Mixture of Gaussians.}  One option for a
variational prior is to assume the mean-field parameters $\mblambda$
are drawn from a mixture of Gaussians.  Let $K$ be the number
of components, $\mbpi$ be a probability vector, $\mbmu_k$, and
$\Sigma_k$ be the parameters of a $d$-dimensional multivariate
Gaussian. The variational prior is
\begin{align*}
  q(\mblambda ; \mbtheta) = \sum_{i=1}^K \mbpi_k \textrm{N}(\mbmu_k, \Sigma_k).
\end{align*}
The parameters $\mbtheta$ contain the probability vector $\mbpi$ as
well as the component means $\mbmu_k$ and variances $\Sigma_k$.  The
mixture locations $\mbmu_k$ capture relationships between different
latent variables. For example, a two-component mixture with two latent
variables (and a mean field variational likelihood) can capture that
the latent variables are either very positive or very negative.

Mixtures can approximate arbitrary distributions (given
enough components), and have been considered as
variational families~\citep{Jaakkola:1998, Lawrence:2000,
  Gershman:2012, Salimans:2013}. In the traditional setup, however,
the mixtures form the variational appproximation on the
latent variables directly.  Here we use it on the variational
parameters; this lets us use a mixture of Gaussians in
many models, including those with discrete latent variables.

\parhead{Variational prior: Normalizing flows.}
Mixtures offer flexible variational priors.
However, in the algorithms we derive, the number of model
likelihood evaluations scales with the number of mixture components;
this is problematic in high dimensions. Further, in high dimensions
the number of
mixtures components can be impractical.
We seek a prior whose computational complexity does not
scale with its modeling flexibility. This motivates normalizing flows.

Normalizing flows are
variational approximations for probability models with differentiable
densities~\citep{Rezende:2015}. Normalizing flows build a
parameterized probability distribution by transforming a simple random
variable $\mblambda_0$ through a sequence of invertible differentiable
functions $f_1$ to $f_K$. Each function transforms its input, so the distribution of the output is a complex
warping of the original random variable $\mblambda_0$.

We can use normalizing flows as a variational prior. Let $\mblambda_k = f_k \circ ... \circ f_1(\mblambda_0)$; then the flow's
density is
 \begin{align*}
 q(\mblambda ; \theta) = q(\mblambda_0) \prod_{k=1}^K \left|\det\left(\frac{\partial f_k}{\partial \mblambda_k} \right)\right|^{-1}.
 \end{align*}
With the normalizing flow prior, the latent variables become dependent
because their variational parameters are deterministic functions of
the same random variable.
The HVM expands the use of normalizing flows to non-differentiable latent variables, such as
those with discrete, ordinal, and discontinuous support.
In~\mysec{def_experimental}, we use normalizing
flows to better approximate posteriors of discrete latent variables.

\parhead{Other Variational Models.}
Many modeling tools can be brought to bear
on building hierarchical variational models.  For example, copulas
explicitly introduce dependence among $d$ random
variables by using joint distributions on
$d$-dimensional hypercubes~\citep{Nelsen:2006}.  \glspl{HVM} can use
copulas as
priors on either point mass or general mean-field likelihoods.  As
another example, we can replace the mixture model prior with a
factorial mixture~\citep{Ghahramani:1995}.  This leads to a
richer posterior approximation.

\subsection{Related work}
There has been much work on learning posterior dependences.
\citet{Saul:1996b,
ghahramani1997structured} develop structured variational
approximations: they factorize the
variational family across subsets of variables, maintaining
certain dependencies in the model. Unlike \acrshortpl{HVM}, however,
structured approximations require model-specific considerations and
can scale poorly when used with black box methods. For example,
\citet{Mnih:2014} develop a structured approximation for
sigmoid belief networks---their approach is restricted to stochastic
feed forward networks, and the variance of the stochastic gradients
increases with the number of layers.
In general, these families complement the construction of \glspl{HVM},
and can be applied as a variational likelihood.

Within the context of more generic inference,
\citet{Titsias:2014,Rezende:2015, Kucukelbir:2016}
propose rich approximating families in
differentiable probability models.  These methods work well in practice; however, they
are restricted to probability models with densities
differentiable with respect to their latent variables.
For undirected models \citet{Agakov:2004} introduced the auxiliary bound for variational
inference we derive. \citet{Salimans:2015} derive the
same bound, but limit their attention to differentiable probability models and
auxiliary distributions defined by Markov transition kernels. \citet{Maaloe:2016}
study auxiliary distributions for semi-supervised learning with deep
generative models.
\citet{Tran:2015} propose
copulas as a way of learning dependencies in factorized
approximations. Copulas can be efficiently extended to
\acrshortpl{HVM}, whereas the full rank approach taken in
\citet{Tran:2015} requires computation quadratic in the number of
latent variables. \citet{Giordano:2015} use linear response theory to
recover covariances from mean-field estimates. Their approach requires
recovering the correct first order moments by mean-field inference
and only provides estimates of smooth functions.

These generic methods can also be building blocks for
\acrshortpl{HVM}, employed as variational priors for
arbitrary mean-field factors.  As in our example with a normalizing
flow prior, this extends their scope to perform inference in discrete
models (and, more generally, non-differentiable models).
In other work, we use Gaussian processes~\citep{Tran:2016}.

\section{Optimizing HVMs}
\label{sec:optimization}

We derive a black box variational inference algorithm for a large class of probability models and
using any hierarchical variational model as the posterior
approximation. Our algorithm enables efficient inference
by preserving both the computational complexity
and variance properties of the stochastic gradients of the variational likelihood.

\label{sec:optimizing}

\paragraph{Hierarchical \glsunset{ELBO}\gls{ELBO}.}
We optimize over the parameters $\mbtheta$ of the variational prior
to find
the optimal distribution within the class of \acrlongpl{HVM}. Using
the \acrshort{HVM}, the \gls{ELBO} is
\begin{align}
\cL(\mbtheta) = \E_{\qhvm(\mbz ; \mbtheta)}[\log p(\mbx, \mbz) - \log \qhvm(\mbz ; \mbtheta)].
\label{eq:marginal-elbo}
\end{align}
The expectation of the first term is tractable as long as we can sample from $q$
and it has proper support. The expectation of the second term is the entropy.
It contains an integral~(\myeqp{mixed-mf}) with respect to the
variational prior, which is analytically intractable in general.

We construct a bound on the entropy. We
introduce a distribution, $r(\mblambda\g\mbz ; \mbphi)$
with parameters $\mbphi$ and apply the variational principle;
\begin{align}
-\E&_\qhvm [\log \qhvm(\mbz)]
\label{eq:entropy-bound}
\\
&\ge -\E_{q(\mbz, \mblambda)}[\log q(\mblambda)+\log q(\mbz \g \mblambda) - \log r(\mblambda \g \mbz ; \mbphi)].
\nonumber
\end{align}
As in variational inference, the bound
in \myeqp{entropy-bound} is exact when $r(\mblambda\g\mbz; \mbphi)$
matches the \emph{variational posterior} $q(\mblambda\g\mbz; \mbtheta)$. From
this perspective, we can view $r$ as a {recursive variational approximation}.
It is a model for the {posterior} $q$ of the mean-field parameters
$\mblambda$ given a realization of the latent variables $\mbz$.

The bound is derived by introducing a term $\KL(q\|r)$. Due to the
asymmetry of KL-divergence, we can also substitute $r$ into
the first rather than the second argument of the KL divergence; this produces
an alternative bound to~\myeqp{entropy-bound}. We can also extend the bound to multi-level \acrlongpl{HVM}, where now we model the posterior distribution $q$ of the
higher levels using higher levels in $r$. More details are available in the appendix.

Substituting the entropy bound (\myeqp{entropy-bound}) into the \gls{ELBO}
(\myeqp{marginal-elbo}) gives a tractable lower bound.
The \emph{hierarchical \gls{ELBO}} is
\begin{align}
\begin{split}
\widetilde{\cL}(&\mbtheta, \mbphi)
= \E_{q(\mbz,\mblambda;\mbtheta)}\Big[\log p (\mbx, \mbz) + \log r(\mblambda \g \mbz  ; \mbphi)
\\&
- \sum_{i=1}^d \log q(z_i \g \mblambda_i) - \log q(\mblambda ;
\mbtheta)\Big].
\end{split}
\label{eq:full-objective}
\end{align}
The hierarchical \gls{ELBO} is tractable, as all of the terms are
tractable. We jointly fit $q$ and $r$ by
maximizing~\myeqp{full-objective} with respect to $\mbtheta$ and
$\mbphi$. Alternatively, the joint maximization can be interpreted as variational EM on
an expanded probability model,
$r(\mblambda\g\mbz;\mbphi)p(\mbz\g\mbx)$. In this light, $\mbphi$ are model parameters and $\mbtheta$ are variational parameters.
Optimizing $\mbtheta$ improves the posterior approximation; optimizing $\mbphi$ tightens the bound on the KL divergence by improving the recursive variational approximation.

We can also analyze \myeqp{full-objective} by rewriting it in terms of
the mean-field \gls{ELBO},
\begin{align*}
\widetilde\cL(\mbtheta, \mbphi) =
\E_q[\cL_{\textsc{mf}}(\mblambda)] + \E_q[\log r(\mblambda \g \mbz ; \mbphi) - \log q(\mblambda ; \mbtheta)].
\end{align*}
where
$\cL_{\textsc{mf}}=\mathbb{E}_{q(\mbz\g\mblambda)}[\log p(\mbx,\mbz) -
\log q(\mbz\g\mblambda)]$.
This shows that \myeqp{full-objective} is a sum of two terms: a
Bayesian model average of the \gls{ELBO} of the variational likelihood,
with weights given by the variational prior $q(\mblambda ; \mbtheta)$; and a correction term
that is a function of both the auxiliary distribution $r$ and the variational prior. Since mixtures (i.e., convex combinations) cannot be sharper
than their components,
$r$ must not be independent of $\mbz$ in order for this bound to be better than the original bound.

\paragraph{Stochastic Gradient of the \gls{ELBO}.}
Before we describe how to optimize the hierarchical \gls{ELBO}, we describe
two types of stochastic gradients of the \gls{ELBO}.

The score function estimator for the \gls{ELBO} gradient applies to both
discrete and continuous latent variable models.
Let $V$ be the score function,
$
V = \nabla_\mblambda \log q(\mbz \g \mblambda).
$
The gradient of the \gls{ELBO} is
\begin{align}
\nabla_{\mblambda}^{score} &\cL
=
\E_{q(\mbz \g \mblambda)}[V (\log p(\mbx, \mbz) - \log q(\mbz \g \mblambda))].
\label{eq:score-grad}
\end{align}
See~\citet{Ranganath:2014} for a derivation. We can construct noisy
gradients from~\myeqp{score-grad} by a Monte Carlo estimate of the
expectation.
In general, the score function estimator exhibits high variance.%
\footnote{%
This is not surprising given that the score function estimator makes
very few restrictions on the class of models, and requires access only
to zero-order information given by the learning signal $\log p - \log
q$.
}
Roughly, the variance of the estimator scales with the number of factors in the learning signal~\citep{Ranganath:2014,Mnih:2014,Rezende:2014}.

In mean-field models, the gradient of the \gls{ELBO} with respect to
$\mblambda_i$ can be separated.
Letting $V_i$ be the local score
$
V_i = \nabla_{\mblambda} \log q(z_i \g \mblambda_i),
$
it is
\begin{align}
 \nabla_{\mblambda_i} \cLmf =
 \mathbb{E}_{q(z_i ; \mblambda_i)}[
 V_i
 (\log p_i(\mbx, \mbz) - \log q(z_i ; \mblambda_i))],
 \label{eq:grad-mf}
\end{align}
where $\log p_i(\mbx, \mbz)$ are the components in the joint distribution
that contain $z_i$. This update is not only local but it also drastically
reduces the variance of~\myeqp{score-grad}. It makes stochastic
optimization practical.

With differentiable latent variables, the estimator
can take advantage of model gradients.
One such estimator uses reparameterization: the \gls{ELBO} is
written in terms of a random variable $\mbepsilon$, whose distribution
$s(\mbepsilon)$ is free of
the variational parameters, and such that $\mbz$ can be written as
a deterministic function $\mbz=\mbz(\mbepsilon;\mblambda)$.
Reparameterization allows gradients of variational parameters to move inside the expectation,
\begin{align*}
\nabla_{\mblambda}^{rep} \cL = \E_{s(\mbepsilon)} [(\nabla_{\mbz} \log p(\mbx, \mbz) -
\nabla_{\mbz} \log q(\mbz)) \nabla_{\mblambda} \mbz(\mbepsilon; \mblambda)].
\end{align*}
The reparameterization gradient constructs noisy gradients from this expression
via Monte Carlo.
Empirically, the reparameterization gradient exhibits lower variance
than the score function gradient~\citep{Titsias:2015}. In the
appendix, we show an analytic equality of the two gradients, which
explains the observed difference in variances.

\paragraph{Stochastic Gradient of the Hierarchical \gls{ELBO}.}
To optimize~\myeqp{full-objective}, we need to compute the
stochastic gradient with respect to the variational hyperparameters
$\mbtheta$ and auxiliary parameters $\mbphi$.
As long as we specify the variational prior $q(\mblambda; \mbtheta)$
to be differentiable, we can apply the reparameterization gradient for
the random variational parameters $\mblambda$. Let $\mbepsilon$ be drawn
from a distribution $s(\mbepsilon)$ such as the standard normal.
Let $\mblambda$
be written as a function of $\mbepsilon$ and $\mbtheta$ denoted
$\mblambda(\mbepsilon ; \mbtheta)$.
The gradient of the hierarchical \gls{ELBO}
with respect to $\mbtheta$ is
\begin{align}
\nabla_\mbtheta & \widetilde{L}(\mbtheta, \mbphi) = \E_{s(\mbepsilon)}[\nabla_\mbtheta \mblambda(\mbepsilon) \nabla_\mblambda \cL_{\textsc{mf}}(\mblambda)]
\nonumber \\
&+
\E_{s(\mbepsilon)}[\nabla_\mbtheta \mblambda(\mbepsilon) \nabla_{\mblambda} [\log r(\mblambda \g \mbz ; \mbphi) - \log q(\mblambda ; \mbtheta)]]
\nonumber \\
&+ \E_{s(\mbepsilon)}[\nabla_\mbtheta \mblambda(\mbepsilon) \E_{q(\mbz \g \mblambda)}[V \log r(\mblambda \g \mbz ; \mbphi)]].
\label{eq:first-gradient}
\end{align}
The first term is the gradient of the original variational approximation scaled
by the chain rule from the reparameterization.
Thus, \acrlongpl{HVM} inherit properties from the original variational approximation
such as the variance
reduced gradient~(\myeqp{grad-mf}) from the mean-field factorization. The second
and third terms try to match $r$ and $q$. The second term is strictly
based on reparameterization, thus it exhibits low variance.

The third term
potentially involves a high variance gradient due to
the appearance of all the latent variables in its gradient. Since the distribution
$q(\mbz \g \mblambda(\mbepsilon; \mbtheta))$ factorizes (by
definition) we can apply
the same variance reduction for $r$ as for the mean-field
model. We examine this below.

\paragraph{Local Learning with $r$.}
The practicality of \acrshortpl{HVM} hinges on
the variance of the stochastic gradients during optimization.
Specifically, any additional variance introduced by $r$ should be minimal.
Let $r_i$ be the terms $\log r(\mblambda \g z_i)$ containing
$z_i$.
Then the last term in \myeqp{first-gradient} can be rewritten as
\begin{align*}
 &\E_{s(\mbepsilon)}[\nabla_\mbtheta \mblambda(\mbepsilon; \mbtheta)
 \E_{q(\mbz \g \mblambda)}[V \log r(\mblambda \g \mbz ; \mbphi)]]
 \\
 &=\E_{s(\mbepsilon)}\left[\nabla_\mbtheta \mblambda(\mbepsilon; \mbtheta) \E_{q(\mbz \g \mblambda)}\left[\sum_{i=1}^d V_i \log r_i(\mblambda \g \mbz ; \mbphi) \right] \right].
\end{align*}
We derive this expression (along with \myeqp{first-gradient}) in the appendix.
When $r_i$ does not depend on many variables, this gradient
combines the computational efficiency of the mean-field with
reparameterization, enabling fast inference for discrete
and continuous latent variable models. This gradient also gives us the
criteria for
building an $r$ that admits efficient stochastic gradients:
$r$ should be differentiable with respect to $\mblambda$,
flexible enough to model the variational posterior
$q(\mblambda\g\mbz)$, and factorize with respect to its dependence on
each $z_i$.

One way to satisfy these criteria is by defining $r$ to be a deterministic
transformation of a factorized distribution. That is, let $\mblambda$ be
the deterministic transform of $\mblambda_0$, and
\begin{align}
r(\mblambda_0 \g \mbz) = \prod_{i=1}^d r({\mblambda_0}_i \g z_i).
\label{eq:r-base-dist}
\end{align}
Similar to normalizing flows, the deterministic transformation from $\mblambda_0$
to $\mblambda$ can be a sequence of invertible, differentiable
functions $g_1$ to $g_k$. However unlike normalizing flows, we let the
inverse functions $g^{-1}$ have a known
parametric form. We call this the \emph{inverse flow}. Under this transformation,
the log density of $r$ is
\begin{align*}
\log r(\mblambda\mid\mbz) &= \log r(\mblambda_0 \g \mbz) + \sum_{k=1}^K \log\left(\left|\det(\frac{\partial g_k^{-1}}{\partial \mblambda_k})\right|\right).
\end{align*}
The distribution $r$ is parameterized by a deterministic transformation of a
factorized distribution.
We can quickly compute the sequence of intermediary $\mblambda$ by applying
the known inverse functions---this enables us to quickly evaluate the log density of inverse
flows at arbitrary points. This contrasts
normalizing flows, where evaluating the log density of a value (not
generated by the flow) requires inversions for each transformation.

This $r$ meets our criteria. It is differentiable, flexible,
and isolates each individual latent variable in a single term.
It maintains the locality of the mean-field inference and is
therefore crucial to the stochastic optimization.

\paragraph{Optimizing the Hierarchical \gls{ELBO} with respect to $\mbphi$.}
We derived how to optimize with respect to $\mbtheta$. Optimizing with
respect to the auxiliary parameters $\mbphi$ is simple. The expectation in the hierarchical \gls{ELBO} (\myeqp{full-objective}) does not depend on $\mbphi$;
therefore we can simply pass the gradient operator inside,
\begin{align}
\nabla_{\mbphi} \widetilde \cL = \E_{q(\mbz,\mblambda)} [\nabla_{\mbphi} \log r(\mblambda \g \mbz, \mbphi)].
\label{eq:r-grad}
\end{align}

\paragraph{Algorithm.}
Algorithm~\ref{alg:comp_vi} outlines the inference procedure, where we evaluate noisy estimates of both gradients using samples
from the joint $q(\mbz,\mblambda)$.  In general, we can compute these gradients
via automatic differentiation systems such as those available in Stan and Theano
\citep{stan-software:2015, Bergstra:2010}. This removes the need for
model-specific computations (note that no assumption has been
made on $\log p(\mbx,\mbz)$ other than the ability to calculate it).

\begin{algorithm}[t]
\SetKwInOut{Input}{Input}
\SetKwInOut{Output}{Output}
 \Input{Model $\log p(\mbx, \mbz)$, \newline Variational model $q(\mbz \g \mblambda) q(\mblambda ; \mbtheta)$}
 \Output{Variational Parameters: $\mbtheta$}
 Initialize $\mbphi$ and $\mblambda$ randomly.

 \While{not converged}{
  Compute unbiased estimate of $\nabla_{\mbtheta}\widetilde{\cL}$.
  \quad (\myeqp{first-gradient})
  Compute unbiased estimate of $\nabla_{\mbphi}\widetilde{\cL}$. \quad
  (\myeqp{r-grad})
  Update $\mbphi$ and $\mblambda$ using stochastic gradient ascent.
 }
 \caption{Black box inference with an \acrshort{HVM}}
 \label{alg:comp_vi}
\end{algorithm}

\mytable{bbi} outlines variational methods and their complexity requirements.
\glspl{HVM}, with a normalizing flow prior, have
complexity linear in the number of latent variables, and the
complexity is proportional
to the length of the flow used to represent $q$ and the inverse flow $r$.

\begin{table*}[!htb]
\centering
\begin{tabular}{lllll}
\toprule
Black box methods & Computation & Storage & Dependency & Class of models
\\
\midrule
\acrshort{BBVI} {\small \citep{Ranganath:2014}} & $\mathcal{O}(d)$ & $\mathcal{O}(d)$ & \xmark &
discrete/continuous\\
\acrshort{DSVI} {\small \citep{Titsias:2014}}& $\mathcal{O}(d^2)$ & $\mathcal{O}(d^2)$ &
\cmark & continuous-diff.\\
\acrshort{CVI} {\small \citep{Tran:2015}} & $\mathcal{O}(d^2)$ & $\mathcal{O}(d^2)$ &
\cmark & discrete/continuous\\
\acrshort{MIXTURE} {\small \citep{Jaakkola:1998}} & $\mathcal{O}(Kd)$ & $\mathcal{O}(Kd)$ & \cmark & discrete/continuous\\
\acrshort{NF} {\small \citep{Rezende:2015}} & $\mathcal{O}(Kd)$ & $\mathcal{O}(Kd)$ & \cmark &
continuous-diff.\\
\acrshort{HVM} w/ \acrshort{NF} prior & $\mathcal{O}(Kd)$ & $\mathcal{O}(Kd)$ & \cmark & discrete/continuous\\
\bottomrule
\end{tabular}
\caption{%
A summary of black box inference methods, which can support either
continuous-differentiable distributions or
both discrete and continuous. $d$ is the number
of latent variables;
for \acrshort{MIXTURE}, $K$ is the number of mixture components;
for \acrshort{NF} procedures, $K$ is the number of transformations.
\vspace{-2ex}
}
\label{table:bbi}
\end{table*}

\Acrlongpl{HVM} with multiple layers can contain
both discrete and differentiable latent variables. Higher
level differentiable variables follow directly from our derivation above.
(See the appendix.)

\paragraph{Inference Networks.}
Classically, variational inference on models
with latent variables associated with a data point
requires optimizing over
variational parameters whose number grows with the size of data. This process can be computationally prohibitive, especially
at test time. Inference
networks~\citep{Dayan:2000, Stuhlmuller:2013, Kingma:2014, Rezende:2014} amortize the cost of estimating these local variational
parameters by tying them together through a neural network. Specifically,
the data-point specific variational parameters are outputs of a neural
network with the data point as input. The parameters of the neural network
then become the variational parameters; this reduces the cost of
estimating the parameters of all the data points to estimating parameters of the inference
network. Inference networks can be applied to~\acrshortpl{HVM}
by making both the parameters of the variational model and
recursive posterior approximation functions of their conditioning sets.

\section{Empirical Study}

We introduced a new class of variational
families and developed efficient black box algorithms for their
computation. We consider a simulated study on a
two-dimensional discrete posterior; we also evaluate our proposed
variational models on deep exponential
families~\citep{Ranganath:2015}, a class of deep generative models
which achieve state-of-the-art results on text analysis. In total, we
train 2 variational models for the simulated study and 12 models over two
datasets.%
\footnote{
An implementation of \glspl{HVM} is available in
Edward~\citep{tran2016edward}, a Python library for probabilistic
modeling.
}

\subsection{Correlated Discrete Latent Variables}
\begin{figure}[tb]
\begin{subfigure}[t]{0.35\columnwidth}
  \centering
  \includegraphics[width=\columnwidth]{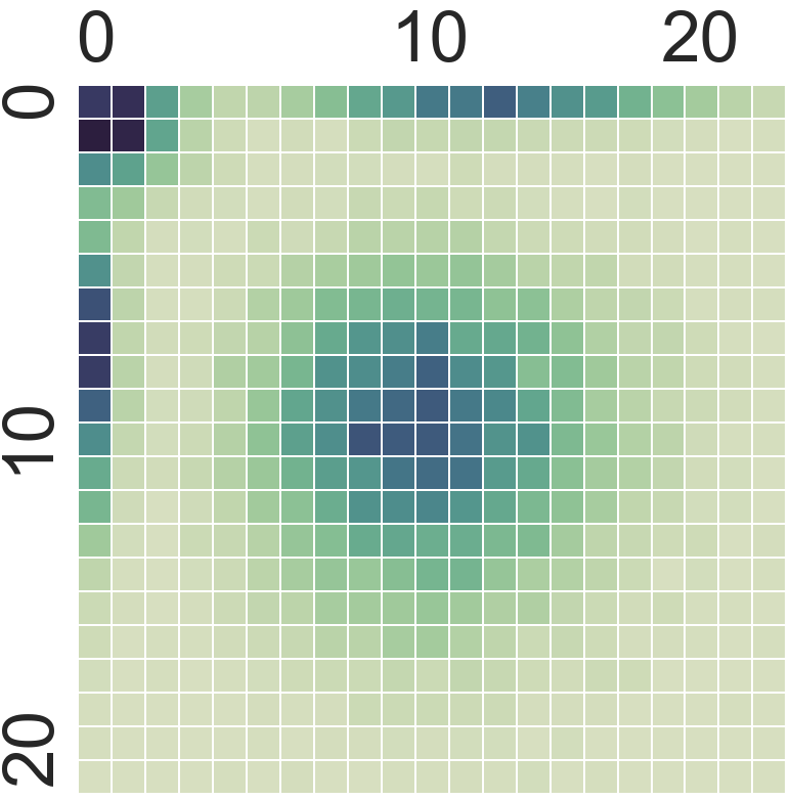}
  \vspace{-2ex}
  \label{sub:true}
\end{subfigure}
\hspace{-1.0em}
\begin{subfigure}[t]{0.35\columnwidth}
  \centering
\includegraphics[width=\columnwidth]{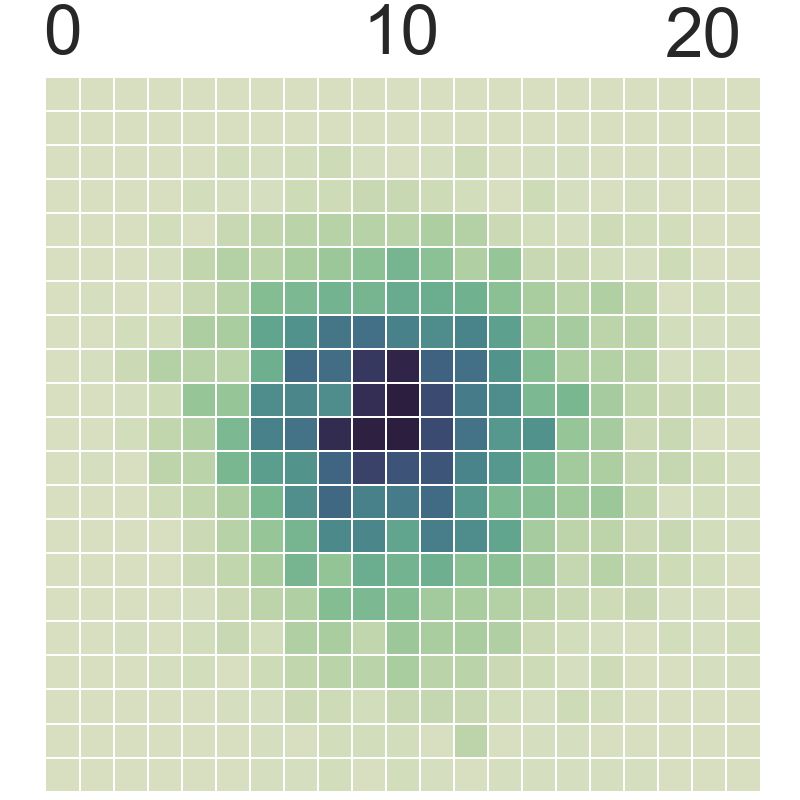}
  \vspace{-2ex}
  \label{sub:mf}
\end{subfigure}
\hspace{-1.0em}
\begin{subfigure}[t]{0.35\columnwidth}
  \centering
\includegraphics[width=\columnwidth]{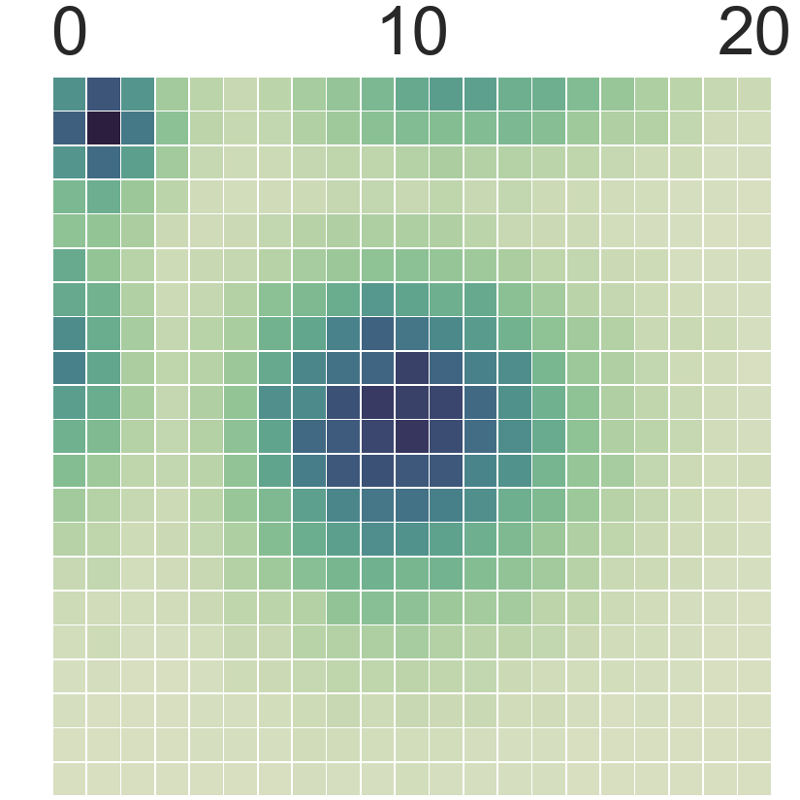}
  \vspace{-2ex}
  \label{sub:mog}
\end{subfigure}
\caption{%
\textbf{(a)} The true posterior, which has correlated latent
variables with countably infinite discrete support.
\textbf{(b)}
Mean-field Poisson approximation.
\textbf{(c)}
\Acrlong{HVM} with a mixture of Gaussians prior.
Using this prior, the \gls{HVM} exhibits
high fidelity to the posterior as it capture multimodality on
discrete surfaces.
}
\label{fig:poisson}
\vskip -0.1in
\end{figure}

Consider a model whose posterior distribution is a pair of discrete
latent variables defined on the countable support
$\{0,1,2,\ldots,\}\times \{0,1,2,\ldots,\}$; \myfig{poisson} depicts its probability
mass in each dimension. The
latent variables are correlated and form a complex multimodal structure.
A mean-field Poisson approximation has difficulty capturing this
distribution; it focuses entirely on the center mass. This contrasts \acrlongpl{HVM},
where we place a mixture prior on the Poisson distributions' rate parameters
 (reparameterized to share the same
support).
This \gls{HVM} fits the various modes of the correlated Poisson latent
variable model and exhibits a ``smoother'' surface
due to its multimodality.

\subsection{Deep Exponential Families}
\label{sec:def_experimental}
Deep exponential families (\glsunset{DEF}\glspl{DEF})~\citep{Ranganath:2015} build
a set of probability models from exponential
families~\citep{Brown:1986}, whose latent structure mimic
the architectures used in deep neural networks.

\paragraph{Model.}
Exponential families are parameterized by a set of natural parameters. We
denote a draw from an unspecified exponential family with natural parameter
$\eta$ as $\expfam(\eta)$. The natural parameter in deep exponential
families are constructed from an inner product of the previous layer with
weights, passed through a link function $g(\cdot)$.

Let $L$ be the total number of layers, $\mbz_{\ell}$ be a vector of latent
variables for layer $\ell$ (with $\mbz_{\ell, k}$ as an element),
and $\mbW_{\ell, k}$ be shared
weights across observations. \glspl{DEF} use
weights with priors, $\mbW_{\ell, k} \sim \expfam_W(\xi)$,
and a prior at the top layer,
$\mbz_{L,k} \sim \expfam_L(\eta)$. The generative process cascades:
for each element $k$ in layer $\ell=L-1,\ldots,1$,
\begin{align*}
\mbz_{\ell, k} &\sim \expfam_\ell(g_\ell(\mbW_{\ell, k}^\top \mbz_{\ell+1}))
\\
\mbx &\sim \textrm{Poisson}(\mbW_{0} \mbz_{1}).
\end{align*}
We model count data with a Poisson likelihood on $\mbx$.
We focus on \glspl{DEF} with discrete latent variables.

The canonical example of a discrete \gls{DEF} is the
\gls{SBN}~\citep{Neal:1990}. The
\gls{SBN} is a Bernoulli \gls{DEF}, with $\mbz_{\ell,k}\in\{0,1\}$. The
other family of models we consider is the Poisson \gls{DEF}, with
\begin{align*}
p(\mbz_{\ell, k} \g \mbz_{l+1},  \mbW_{l, k}) \sim
\textrm{Poisson}(\log (1 + \exp(\mbz_{l+1}^\top \mbW_{l, k}))),
\end{align*}
for each element $k$ in the layer $\ell$.
In the \gls{SBN}, each observation either turns a feature on or off.
In a Poisson \gls{DEF}, each observation counts each feature a
positive integer number of times. This means Poisson \glspl{DEF} are a
multi-feature generalization of \glspl{SBN}.

\paragraph{Variational Models.}
We consider the variational approximation that adds dependence
to the $\mbz's$. We parameterize each variational prior $q(\mblambda_{\mbz_i})$ with a normalizing flow of length $2$, and use the inverse flow of
length $10$ for $r(\mblambda_{\mbz_i})$.
We use planar transformations~\citep{Rezende:2015}. In a pilot study, we
found little improvement with longer flow lengths.
We compare to the mean-field approximation from~\citet{Ranganath:2015} which
achieves state of the art results on text.

\paragraph{Data and Evaluation.}
We consider two text corpora of news and scientific articles---
\emph{The New York Times} (NYT) and \emph{Science}. Both have 11K documents. NYT
consists of 8K terms and \emph{Science} consists of 5.9K terms. We
train six models for each data set.

We examine held out perplexity following the same criteria as
\citet{Ranganath:2015}.
This is a document complete evaluation metric~\citep{Wallach:2009a} where the words are tested independently.
As our evaluation uses data not included in posterior inference,
it is possible for the mean-field family to outperform \acrshortpl{HVM}.

\paragraph{Results.}
\begin{table}[!tb]
\centering
\begin{tabular}{llllll}
\toprule
& Model &\acrshort{HVM} & Mean-Field
\\
\midrule
\textbf{Poisson} & 100 & {\bf 3386} & 3387\\
& 100-30 & {\bf 3396} & 3896\\
& 100-30-15 & {\bf 3346} & 3962\\
\midrule
\textbf{Bernoulli} & 100 & {\bf 3060} & 3084 \\
& 100-30 & 3394 & {\bf 3339} \\
& 100-30-15 & \textbf{3420} & 3575 \\
\bottomrule
\end{tabular}
\caption{%
\textit{New York Times}. Held-out perplexity (lower is better).
\Acrlongpl{HVM} outperform mean-field in five
models. Mean-field~\citep{Ranganath:2015} fails at multi-level Poissons;
\acrshortpl{HVM} make it possible to study multi-level Poissons.
}
\label{table:def:nyt}
\end{table}

\begin{table}[tb]
\centering
\begin{tabular}{llll}
\toprule
& Model &\acrshort{HVM} & Mean-Field
\\
\midrule
\textbf{Poisson} & 100 & \textbf{3327} & 3392\\
& 100-30 & \textbf{2977} & 3320\\
& 100-30-15 & \textbf{3007} & 3332\\
\midrule
\textbf{Bernoulli} & 100 & \textbf{3165} & 3166\\
& 100-30 &  \textbf{3135} & 3195 \\
& 100-30-15 & \textbf{3050}  & 3185 \\
\bottomrule
\end{tabular}
\caption{%
\textit{Science}. Held-out perplexity (lower is better). \acrshort{HVM} outperforms mean-field on all six models. \Acrlongpl{HVM} identify that multi-level Poisson models are best, while mean-field does not.
\vspace{-3ex}
}
\label{table:def:science}
\end{table}

\glsunset{HVM}
\glspl{HVM} achieve better performance over
six models and two datasets, with a mean improvement in perplexity
of 180 points. (Mean-field works better on only the
two layer Bernoulli model on NYT.) From
a data modeling viewpoint, we find that for \emph{The New York Times} there is little advantage to multi-layer models, while on \emph{Science}
multi-layer models outperform their single layer counterparts.
Overall, \acrlongpl{HVM} are less sensitive to inference in multi-layer
models,
as evidenced by the generally lower performance of mean-field with multiple layers.
\glspl{HVM} make it feasible to work with multi-level Poisson
models. This is particularly important on \emph{Science}, where
\acrlongpl{HVM} identifies that multi-level Poisson models are best.

\section{Discussion}

We present \acrlongpl{HVM}, a
rich class of posterior approximations constructed by placing
priors on existing variational families.
These priors encapsulate different modeling assumptions of the
posterior and we explore several choices.
We develop a black box algorithm
can fit any \acrshort{HVM}.
There are several avenues for future work:
studying alternative entropy bounds; analyzing \acrshortpl{HVM} in the
empirical Bayes framework;
and using other data modeling tools to build
new variational models.

\paragraph{Acknowledgements.}
This work is supported by NSF IIS-0745520, IIS-1247664, IIS-1009542, ONR N00014-11-1-0651, DARPA FA8750-14-2-0009, N66001-15-C-4032, Facebook, Adobe, Amazon, NVIDIA, the Porter Ogden Jacobus
Fellowship, the Seibel Foundation, and John Templeton Foundations.

\clearpage
\section*{References}
\renewcommand{\bibsection}{}
\bibliography{bib.bib}
\bibliographystyle{apalike}

\clearpage
\appendix
\section{Appendix}
\paragraph{Relationship to empirical Bayes and RL.}
The augmentation with a variational prior has strong ties to empirical Bayesian methods,
which use data to estimate hyperparameters of a prior distribution
\citep{Robbins:1964,Efron:1973}. In general, empirical Bayes considers the fully
Bayesian treatment of a hyperprior on the original prior---here, the variational prior on the original mean-field---and proceeds to integrate it out. As this
is analytically intractable, much work has been on parametric
estimation, which seek point estimates rather than the whole distribution
encoded by the hyperprior. We avoid this at the level of the hyperprior (variational prior) via the hierarchical ELBO; however, our procedure can
be viewed in this framework at one higher level. That is, we seek a point
estimate of the "variational hyperprior" which governs the parameters on the
variational prior.

A similar methodology also arises in the policy search literature
\citep{Ruckstiess:2008,Sehnke:2008}. Policy search methods aim to maximize the
expected reward for a sequential decision-making task, by positing a
distribution over trajectories and proceeding to learn its parameters. This
distribution is known as the policy, and an upper-level policy considers a
distribution over the original policy. This encourages exploration in the
latent variable space and can be seen as a form of annealing.

\paragraph{Tractable bound on the entropy.}
Deriving an analytic expression for the entropy of $\qhvm$ is generally intractable due
to the integral in the definition of $\qhvm$.
However, it is tractable when we know the distribution $q(\mblambda \g \mbz)$. This can be seen by noting from standard Bayes' rule that
\begin{align}
q(\mbz) q(\mblambda \g \mbz) = q(\mblambda) q(\mbz \g \mblambda),
\label{eq:total-q}
\end{align}
and that the right hand side is specified by the construction of the hierarchical variational model.
Note also that $q(\mblambda\g\mbz)$ can be interpreted as the posterior distribution
of the original variational parameters $\mblambda$ given the model, thus we will denote it as $\qpost(\mblambda\g\mbz)$.

In general, computing $\qpost(\mblambda \g \mbz)$ from the specification
of the hierarchical variational model is as hard as the integral
needed to compute the entropy.
Instead, we approximate $\qpost$ with
an auxiliary distribution $r(\mblambda \g \mbz ; \mbphi)$ parameterized
by $\mbphi$. This yields a bound on the entropy in terms of the analytically
known distributions $r(\mblambda\g\mbz)$, $q(\mbz \g \mblambda)$, and $q(\mblambda)$.

First note that the KL-divergence between two distributions is greater than
zero, and is precisely zero only when the two distributions are equal. This
means the entropy can be bounded as follows:
\begin{align*}
-\E&_\qhvm [\log \qhvm(\mbz)]
\\
&= -\E_\qhvm[\log \qhvm(\mbz) - \KL(\qpost(\mblambda \g \mbz) || \qpost(\mblambda \g \mbz))]
\\
& \geq -\E_\qhvm[\log \qhvm(\mbz) + \KL(\qpost(\mblambda \g \mbz) || r(\mblambda \g \mbz ; \mbphi))]]
\\
&= -\E_\qhvm[\E_\qpost[ \log \qhvm(\mbz)+ \log \qpost(\mblambda \g \mbz)
\\
&\quad - \log r(\mblambda \g \mbz ; \mbphi)]]
\\
&= -\E_{q(\mbz, \mblambda)}[\log \qhvm(\mbz)+\log \qpost(\mblambda \g \mbz) - \log r(\mblambda \g \mbz ; \mbphi)].
\end{align*}
Then by \myeqp{total-q}, the bound simplifies to
\begin{align*}
-\E&_\qhvm [\log \qhvm(\mbz)]
\\
&\ge -\E_{q(\mbz, \mblambda)}[\log q(\mblambda)+\log q(\mbz \g \mblambda) - \log r(\mblambda \g \mbz ; \mbphi)].
\nonumber
\end{align*}
A similar bound in derived by~\citet{Salimans:2015} directly for $\log p(x)$.

In the above derivation, the approximation $r$ to the variational posterior
$\qpost(\mblambda \g \mbz)$ is placed as the second argument of a KL-divergence
term. Replacing the first argument instead yields a different tractable upper
bound as well.
\begin{align*}
-\E&_\qhvm [\log q(\mbz)]
\nonumber \\
&= \E_\qhvm[-\log q(\mbz) + \KL(\qpost(\mblambda \g \mbz) || \qpost(\mblambda \g \mbz))]
\nonumber \\
& \leq \E_\qhvm[-\log q(\mbz) + \KL(r(\mblambda \g \mbz ; \mbphi) || \qpost(\mblambda \g \mbz))]]
\nonumber \\
&= \E_\qhvm[\E_r[-\log q(\mbz) - \log \qpost(\mblambda \g \mbz) + \log r(\mblambda \g \mbz ; \mbphi)]]
\nonumber \\
&= \E_\qhvm[\E_r[-\log q(\mbz) - \log \frac{q(\mbz \g \mblambda) q(\mblambda)}{q(\mbz)} + \log r(\mblambda \g \mbz ; \mbphi)]]
\nonumber \\
&= \E_\qhvm[\E_r[-\log q(\mblambda) - \log q(\mbz \g \mblambda) + \log r(\mblambda \g \mbz ; \mbphi)]].
\end{align*}
The bound is also tractable when $r$ and $\qhvm$ can be sampled and all distributions are analytic.
The derivation of these two bounds parallels the development of expectation
propagation~\citep{Minka:2001b} and variational Bayes~\citep{Jordan:1999a}
which are based on alternative forms of the KL-divergence\footnote{Note that
the first bound, which corresponds to the objective in expectation
propagation (EP), directly minimizes $\KL(q\|r)$ whereas EP only minimizes
this locally.}. Exploring the role
and relative merits of both bounds we derive in the context of variational
models will be an important direction in the study of variational models with
latent variables.

The entropy bound is tighter than the trivial
conditional entropy bound of $\mathbb{H}[\qhvm] \geq \mathbb{H}[q \g
\mblambda]$~\citep{Cover:2012}.
This bound is attained when specifying the recursive
approximation to be the prior; i.e., it is the special case when
$r (\mblambda \g \mbz;\mbphi) = q(\mblambda; \mbtheta)$.

\paragraph{Gradient Derivation.}
We derive the gradient of the hierarchical \gls{ELBO} using its mean-field representation:
\begin{align*}
\widetilde\cL(\mbtheta, \mbphi) = \E_q[\cL(\mblambda)] + \E_q[(\log r(\mblambda
\g \mbz ; \mbphi) - \log q(\mblambda ; \mbtheta))].
\end{align*}
Using the reparameterization $\mblambda(\mbepsilon ; \mbtheta)$, where $\mbepsilon \sim s$, this
is
\begin{align*}
\widetilde\cL &(\mbtheta, \mbphi) = \E_{s(\mbepsilon)}[\cL(\mblambda(\mbepsilon ; \mbtheta))]
\\
&+ \E_{s(\mbepsilon)}[ \E_{q(\mbz \g \mblambda)}[(\log r(\mblambda(\mbepsilon ; \mbtheta) \g \mbz ; \mbphi)] - \log q(\mblambda(\mbepsilon ; \mbtheta) ; \mbtheta))].
\end{align*}
We now differentiate the three additive terms with respect to $\mbtheta$. As in
the main text, we suppress $\mbtheta$ in the definition of $\mblambda$ when clear and
define the score function
\begin{align*}
V = \nabla_\mblambda \log q(\mbz \g \mblambda).
\end{align*}
By the chain rule the derivative of the first term is
\begin{align*}
\nabla_\mbtheta \E_{s(\mbepsilon)}[\cL(\mblambda(\mbepsilon ; \mbtheta))] = \E_{s(\mbepsilon)}[\nabla_\mbtheta \mblambda(\mbepsilon) \nabla_\mblambda \cL(\mblambda)].
\end{align*}
We now differentiate the second term:
\begin{align*}
& \nabla_\mbtheta \E_{s(\mbepsilon)}[\E_{q(\mbz \g \mblambda)}[\log r(\mblambda(\mbepsilon ; \mbtheta) \g \mbz ; \mbphi]]
\\
&=\nabla_\mbtheta \E_{s(\mbepsilon)}\left[\int q(\mbz \g \mblambda) \log r(\mblambda(\mbepsilon ; \mbtheta) \g \mbz ; \mbphi)\d\mbz\right]
\\
&=\E_{s(\mbepsilon)}\left[\nabla_\mbtheta\left[\int q(\mbz \g \mblambda) \log r(\mblambda(\mbepsilon ; \mbtheta) \g \mbz ; \mbphi)\d\mbz\right]\right]
\\
&=\E_{s(\mbepsilon)}\left[\nabla_\mbtheta \mblambda(\mbepsilon)
\nabla_\mblambda \left[\int q(\mbz \g \mblambda) \log r(\mblambda(\mbepsilon ;
\mbtheta) \g \mbz ; \mbphi)\d\mbz\right]\right].
\end{align*}
Applying the product rule to the inner derivative gives
\begin{align*}
\nabla_\mblambda &\left[\int q(\mbz \g \mblambda) \log r(\mblambda(\mbepsilon ; \mbtheta) \g \mbz ; \mbphi)\d\mbz\right]
\\
&= \int \nabla_\mblambda q(\mbz \g \mblambda) \log r(\mblambda(\mbepsilon ; \mbtheta) \g \mbz ; \mbphi)\d\mbz
\\
&\quad +\int q(\mbz \g \mblambda) \nabla_\mblambda \log r(\mblambda(\mbepsilon ; \mbtheta) \g \mbz ; \mbphi) \d\mbz
\\
&= \int \nabla_\mblambda \log q(\mbz \g \mblambda) q(\mbz \g \mblambda) \log r(\mblambda(\mbepsilon ; \mbtheta) \g \mbz ; \mbphi)\d\mbz
\\
&\quad +\int q(\mbz \g \mblambda) \nabla_\mblambda \log r(\mblambda(\mbepsilon ; \mbtheta) \g \mbz ; \mbphi) \d\mbz
\\
&= \E_{q(\mbz \g \mblambda)}[V \log r(\mblambda(\mbepsilon ; \mbtheta) \g \mbz ; \mbphi)]
\\
&\quad +\E_{q(\mbz \g \mblambda)}[\nabla_\mblambda \log r(\mblambda(\mbepsilon ; \mbtheta) \g \mbz ; \mbphi)].
\end{align*}
Substituting this back into the previous expression gives the gradient of the second term
\begin{align*}
&\E_{s(\mbepsilon)}[\nabla_\mbtheta \mblambda(\mbepsilon) \E_{q(\mbz \g \mblambda)}[V \log r(\mblambda(\mbepsilon ; \mbtheta) \g \mbz ; \mbphi)]]
\\
&\quad +\E_{s(\mbepsilon)}[\nabla_\mbtheta \mblambda(\mbepsilon) \E_{q(\mbz \g \mblambda)}[\nabla_\mblambda \log r(\mblambda(\mbepsilon ; \mbtheta) \g \mbz ; \mbphi)]]
\end{align*}
The third term also follows by the chain rule
\begin{align*}
\nabla_\mbtheta &\E_{s(\mbepsilon)}[\log q(\mblambda(\mbepsilon ; \mbtheta) ; \mbtheta)] \\
&= \E_{s(\mbepsilon)}[\nabla_\mbtheta \mblambda(\mbepsilon) \nabla_\mblambda \log q(\mblambda ; \mbtheta) + \nabla_\mbtheta \log q(\mblambda ; \mbtheta)]
\\
&= \E_{s(\mbepsilon)}[\nabla_\mbtheta \mblambda(\mbepsilon) \nabla_\mblambda \log q(\mblambda ; \mbtheta)]
\end{align*}
where the last equality follows by
\begin{align*}
\E_{s(\mbepsilon)}[\nabla_\mbtheta \log q(\mblambda ; \mbtheta)]
= \E_{q(\mblambda ; \mbtheta)}[\nabla_\mbtheta \log q(\mblambda ; \mbtheta)]
= \mb{0}.
\end{align*}
Combining these together gives the total expression for the gradient
\begin{align*}
\nabla_\mbtheta & \widetilde{L}(\mbtheta, \mbphi) = \E_{s(\mbepsilon)}[\nabla_\mbtheta \mblambda(\mbepsilon) \nabla_\mblambda \cLmf(\mblambda)]
\nonumber \\
&+
\E_{s(\mbepsilon)}[\nabla_\mbtheta \mblambda(\mbepsilon) \nabla_{\mblambda} [\log r(\mblambda \g \mbz ; \mbphi) - \log q(\mblambda ; \mbtheta)]]
\nonumber \\
&+ \E_{s(\mbepsilon)}[\nabla_\mbtheta \mblambda(\mbepsilon) \E_{q(\mbz \g \mblambda)}[V \log r(\mblambda \g \mbz ; \mbphi)]].
\end{align*}

\paragraph{Introducing $r_i$ to the gradient.}
One term of the gradient involves the product of the score
function with all of $r$,
\begin{align*}
\E_{s(\mbepsilon)}[\nabla_\mbtheta \mblambda(\mbepsilon) \E_{q(\mbz \g \mblambda)}[V \log r(\mblambda \g \mbz ; \mbphi)]].
\end{align*}
Localizing (Rao-Blackwellizing) the inner expectation as in ~\citet{Ranganath:2014,Mnih:2014}
can drastically reduce the variance. Recall that
\begin{align*}
q(\mbz \g \mblambda) = \prod_{i=1}^d q(z_i \g \mblambda_i).
\end{align*}
Next, we define $V_i$ to be the score functions of the factor. That is
\begin{align*}
V_i = \nabla_{\mblambda} \log q(z_i \g \mblambda_i).
\end{align*}
This is a vector with nonzero entries corresponding to $\mblambda_i$.
Substituting the factorization into the gradient term yields
\begin{align}
\E_{s(\mbepsilon)}\left[\nabla_\mbtheta \mblambda(\mbepsilon) \sum_{i=1}^d \E_{q(\mbz \g \mblambda)}[V_i \log r(\mblambda \g \mbz ; \mbphi)] \right].
\label{eq:to-sub}
\end{align}
Now we define $r_i$ to be the terms in $\log r$ containing $z_i$ and $r_{-i}$ to be
the remaining terms. Then the inner expectation in the gradient term is
\begin{align*}
\sum_{i=1}^d &\E_{q(\mbz \g \mblambda)}[V_i (\log r_i(\mblambda \g \mbz ; \mbphi) + \log r_{-i}(\mblambda \g \mbz ; \mbphi))]
\\
&=\sum_{i=1}^d \E_{q(z_{i} \g \mblambda)}[V_i \E_{q(\mbz_{-i} \g \mblambda)} [\log r_i(\mblambda \g \mbz ; \mbphi) + \log r_{-i}(\mblambda \g \mbz ; \mbphi)]],
\\
&= \sum_{i=1}^d \E_{q(\mbz \g \mblambda)}[V_i \log r_i(\mblambda \g \mbz ; \mbphi)],
\end{align*}
where the last equality follows from the expectation of the score function of a distribution is
zero. Substituting this back into~\myeqp{to-sub} yields the desired result
\begin{align*}
 &\E_{s(\mbepsilon)}[\nabla_\mbtheta \mblambda(\mbepsilon; \mbtheta) \E_{q(\mbz \g \mblambda)}[V \log r(\mblambda \g \mbz ; \mbphi)]]
 \\
 &=\E_{s(\mbepsilon)}\left[\nabla_\mbtheta \mblambda(\mbepsilon; \mbtheta) \E_{q(\mbz \g \mblambda)}\left[\sum_{i=1}^d V_i \log r_i(\mblambda \g \mbz ; \mbphi) \right] \right].
\end{align*}

\paragraph{Equality of Two Gradients.}
We now provide a direct connection between the score gradient and the
reparameterization gradient.
We carry this out in one-dimension for clarity, but the same principle holds in
higher dimensions. Let $Q$ be the cumulative distribution function (CDF) of $q$
and let $z = T(\mbz_0;\mblambda)$ be reparameterizable in terms of a uniform random
variable $\mbz_0$ (inverse-CDF sampling).
We focus on the one dimensional case for simplicity.
Recall integration by parts computes a definite integral as
\begin{align*}
\int_{\operatorname{supp}(\mbz)} w(\mbz) &\d v(\mbz) \\
&= |w(\mbz) v(\mbz)|_{\operatorname{supp}(\mbz)} - \int_{\operatorname{supp}(\mbz)} v(\mbz) \d w(\mbz),
\end{align*}
where the $|\cdot|$ notation indicates evaluation of a portion of the integral. In
the subsequent derivation, we let $w(\mbz) = \log p (\mbx, \mbz) - \log q(\mbz)$, and let
$dv(\mbz) = \nabla_\mblambda \log q(\mbz) q(\mbz) =  \nabla_\mblambda q(\mbz)$.

Recall that we assume that we can CDF-transform $\mbz$ and that the transformation is differentiable.
That is, when $\mbu$ is a standard uniform random variable, $\mbz = Q^{-1}(\mbu,
\mblambda)$. Then
\begin{align*}
&\nabla_{\mblambda}^{score} \cL
= \E_{q(\mbz \g \mblambda)}[\nabla_{\mblambda} \log
q(\mbz \g \mblambda) (\log p(\mbx, \mbz) - \log q(\mbz \g \mblambda))]
\\
&\quad = \int_{supp(\mbz)} \nabla_{\mblambda}
q(\mbz \g \mblambda) (\log p(\mbx, \mbz) - \log q(\mbz \g \mblambda))] \d\mbz
\\
&\quad = \left|\left[\int_{\mbz} \nabla_{\mblambda} q(\mbz \g \mblambda)\d\mbz\right] (\log p(\mbx, \mbz) - \log q(\mbz \g \mblambda)) \right|_{supp(\mbz)}
\\
&\qquad - \int \left[\int_{\mbz} \nabla_{\mblambda} q(\mbz \g \mblambda)\d\mbz \right] \nabla_{\mbz}[\log p(\mbx, \mbz) - \log q(\mbz \g \mblambda)] \d\mbz
\\
&\quad = \left| \nabla_{\mblambda} Q(\mbz \g \mblambda) (\log p(\mbx, \mbz) - \log q(\mbz \g \mblambda)) \right|_{supp(\mbz)}
\\
&\qquad - \int \nabla_{\mblambda} \left[ Q(\mbz \g \mblambda) \right] \nabla_{\mbz}[\log p(\mbx, \mbz) - \log q(\mbz \g \mblambda)] \d\mbz
\\
&\quad = \left| \nabla_{\mblambda} Q(\mbz \g \mblambda) (\log p(\mbx, \mbz) - \log q(\mbz \g \mblambda)) \right|_{supp(\mbz)}
\\
&\qquad + \int q(\mbz \g \mblambda) \nabla_{\mblambda} \left[ \mbz \right]
\nabla_{\mbz}[\log p(\mbx, \mbz) - \log q(\mbz \g \mblambda)] \d\mbz
\\
&\quad = \left| \nabla_{\mblambda} Q(\mbz \g \mblambda) (\log p(\mbx, \mbz) - \log q(\mbz \g \mblambda)) \right|_{supp(\mbz)}
\\
&\qquad + \nabla_{\mblambda}^{rep} \cL,
\end{align*}
where the second to last equality follows by the derivative of the CDF function~\citep{Hoffman:2015}.
By looking at the Monte Carlo expression of both sides,
we can see the reduction in variance that the reparameterization gradient has
over the score gradient comes from the analytic computation of the gradient of
the definite integral (which has value $\mb{0}$).

\paragraph{Hyperparameters and Convergence.}
We study one, two, and three layer \glspl{DEF} with 100, 30,
and 15 units respectively and set prior hyperparameters following \citet{Ranganath:2015}.
For \glspl{HVM}, we use Nesterov's
accelerated gradient with momentum parameter of $0.9$, combined with
RMSProp with a scaling factor of $10^{-3}$, to maximize the lower
bound. For the mean-field family, we use the learning rate hyperparameters
from the original authors'. The \glspl{HVM} converge faster on Poisson models
relative to Bernoulli models. The one layer Poisson model was the fastest to infer.

\paragraph{Multi-level $q(\mblambda ; \mbtheta)$ and Optimizing with Discrete Variables in the Variational Prior.}
As mentioned in the main text \Acrlongpl{HVM} with multiple layers can contain
both discrete and differentiable latent variables. Higher
level differentiable variables follow directly from our derivation above.
Discrete variables
in the prior pose a difficulty due to high variance,
as the learning signal contains the entire model. Local expectation
gradients~\citep{Titsias:2015} provide
an efficient gradient estimator for variational approximations
over discrete variables with small support---done by analytically marginalizing
over each discrete variable individually. This approach can
be combined with the gradient in Equation 8 of the main text to form
an efficient gradient estimator.

In the setting where the prior has discrete variables,
optimization requires a little more work. First we note that in a non-degenerate
mean-field setup that
the $\mblambda$'s are differentiable parameters of probability distributions.
This means they will always, conditional on the discrete variables, be
differentiable in the variational prior. This means that we can both compute the
gradients for these parameters using the technique from above and that the
discrete variables exist at a higher level of the hierarchical variational model;
these discrete variables can be added to $r$ conditional on everything else.
The gradients of discrete variables can be computed using the score gradient, but
Monte Carlo estimates of this will have high variance due to no simplification
of the learning signal (like in the mean-field). We can step around this issue
by using local expectation gradients~\citep{Titsias:2015}
which marginalize out one variable at a time to get low variance stochastic
gradients. This is generally tractable when the discrete variables have small
support such as the binary variables in the factorial mixture

\end{document}